\documentclass{article}
\pdfoutput=1

\usepackage[final,nonatbib]{neurips_2024}

\usepackage{graphicx}
\usepackage{amsmath}
\usepackage{amssymb}
\usepackage{mathtools}
\usepackage{caption}
\usepackage{float}
\usepackage{enumitem}
\usepackage{bbm}
\usepackage{wrapfig}
\usepackage{subfigure}
\usepackage{multirow}
\usepackage{comment}

\usepackage[utf8]{inputenc} 
\usepackage[T1]{fontenc}    
\usepackage{hyperref}       
\usepackage{url}            
\usepackage{booktabs}       
\usepackage{amsfonts}       
\usepackage{nicefrac}       
\usepackage{microtype}      
\usepackage{xcolor}         

\usepackage{cleveref}

\DeclareMathOperator*{\argmin}{argmin}
\DeclareMathOperator*{\argmax}{argmax}
\Crefname{equation}{Eq.}{Eqs.}
\Crefname{figure}{Fig.}{Figs.}
\Crefname{table}{Tab.}{Tabs.}
\Crefname{section}{Sec.}{Secs.}

\newcommand{\parag}[1]{\paragraph{#1}}
\newcommand{\Teta}[0]{\mathbf{\Theta}}

\newcommand{\bx}{\mathbf{x}}
\newcommand{\bz}{\mathbf{z}}
\newcommand{\bu}{\mathbf{u}}
\newcommand{\bX}{\mathbf{X}}
\newcommand{\bP}{\mathbf{P}}
\newcommand{\bp}{\mathbf{p}}

\newcommand{\mbf}{\mathbf{f}}

\newcommand{\bV}{\mathbf{V}}

\title{Reconstruction of Manipulated Garment with Guided Deformation Prior}

\author{%
Ren Li \qquad Corentin Dumery \qquad Zhantao Deng \qquad Pascal Fua \\
Computer Vision Lab, EPFL\\
Lausanne, Switzerland \\
{\small \texttt{ren.li@epfl.ch}}\quad {\small \texttt{corentin.dumery@epfl.ch}}\quad {\small \texttt{zhantao.deng@epfl.ch}}\quad {\small \texttt{pascal.fua@epfl.ch}}  \\
}

\begin{document}

\maketitle


\begin{abstract}

Modeling the shape of garments has received much attention, but most existing approaches assume the garments to be worn by someone, which constrains the range of shapes they can assume. In this work, we address shape recovery when garments are being manipulated instead of worn, which gives rise to an even larger range of possible shapes. To this end, we leverage the implicit sewing patterns (ISP) model for garment modeling and extend it by adding a diffusion-based deformation prior to represent these shapes. To recover 3D garment shapes from incomplete 3D point clouds acquired when the garment is folded, we map the points to UV space, in which our priors are learned, to produce partial UV maps, and then fit the priors to recover complete UV maps and 2D to 3D mappings. Experimental results demonstrate the superior reconstruction accuracy of our method compared to previous ones, especially when dealing with large non-rigid deformations arising from the manipulations.

\end{abstract}
 

\section{Introduction}
\label{sec:intro}
Garments play an important role in our daily lives, as we interact with them through wearing, folding, and manipulating them. Therefore, the ability to recover their 3D shape is important in many fields, including virtual try-on, VR/AR, and robotic manipulation. However, since garments are non-rigid thin structures with a near-infinite number of degrees of freedom, accurate reconstruction remains a challenge, especially in the presence of massive self-occlusions caused by folding or crumpling. 

\begin{figure}
    \centering
    \includegraphics[width=0.99\textwidth]{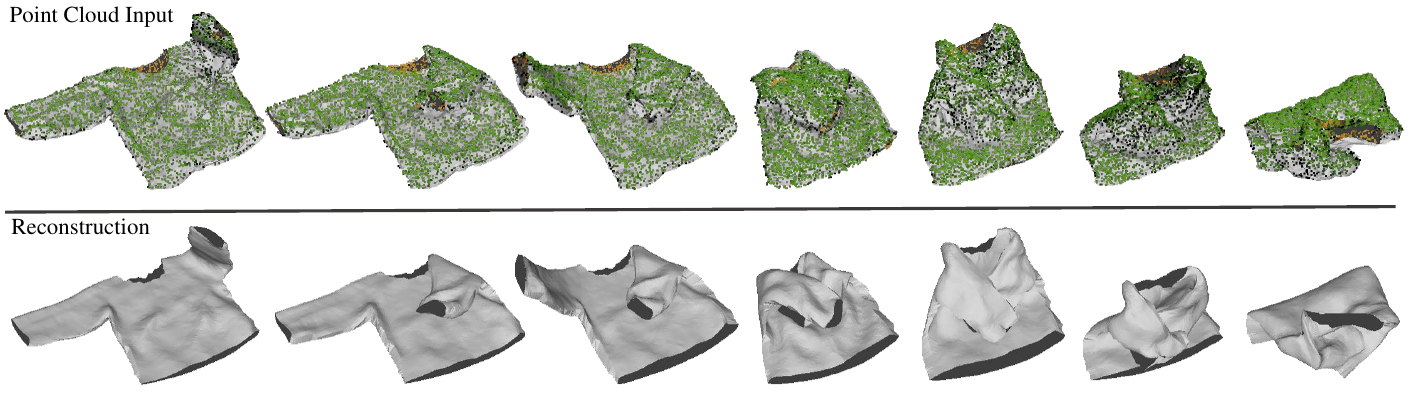}
    \caption{Recovering the 3D shape of folded and crumpled garments from incomplete point clouds. \textbf{Top}: The input point clouds (green) overlaid on the ground truth meshes (gray). \textbf{Bottom}: Our reconstructions.}
    \label{fig:teaser}
\end{figure}

Most existing techniques focus on reconstructing garments being worn by someone and therefore constrained by the body shape. This limits the amount of crumpling and provides a shape prior that can be exploited. In this paper, we address the even more challenging problem of recovering the shape of garments not being worn and therefore possibly assuming arbitrary shapes, such as those of Fig.~\ref{fig:teaser}.

To this end, we start from the  Implicit Sewing Patterns (ISP) model~\cite{Li23a}. As in models used by clothing designers, each garment consists of individual 2D panels. Their 2D shape is defined by a Signed Distance Function and 3D shape by a 2D to 3D mapping. We chose this formalism because it can handle complex  garments with various geometries, while preserving differentiability with respect to observations. However, its 3D parameterization is limited to a single rest state for each garment and is designed to be draped on a human body. 

To handle garments unconstrained by the wearer's body, we introduce a prior to represent the many plausible deformations, including folding and crumpling. It is learned using a generative diffusion model that generates 2D positional UV maps that are applicable to many different garments. We use it to recover 3D garment shapes from incomplete 3D point clouds, such as those that are acquired using a laser scanner when the garment is folded.  

In practice, given that every panel in the ISP model is registered to a unified 2D space parameterized by its UV coordinates, we train a {\it UV mapper} to assign the points to individual panels and to project them into the corresponding UV-space, as shown at the top of Fig.~\ref{fig:pip}. This yields partial UV maps for each panel.  We then fit the 2D panels and use a guided reverse diffusion process to generate complete UV maps from the partial ones.

We validate our approach on the data from the VR-Folding dataset~\cite{Xue23b}, where point clouds are generated from multi-view RGBD images. As shown in Fig. \ref{fig:teaser}, our approach accurately reconstructs 3D garment meshes under high levels of deformation and self-occlusion. We also demonstrate that our algorithm can handle real point cloud data. Notably, our method achieves this without requiring explicit prior knowledge of the garment geometry, further demonstrating its practical applicability. This goes well-beyond prior diffusion-based work~\cite{Guo24} that can only model the deformations of a single specific garment worn on the human body. Our implementation and model weights are available at \url{https://github.com/liren2515/GarmentFolding}.

\section{Related Work}

\parag{Non-Rigid 3D Reconstruction.} 

Reconstructing non-rigid deforming objects has been a longstanding research area in computer vision and graphics. One line of work~\cite{Ngo15b,Perriollat11,Salzmann07c,Yu15a,Pumarola18,Kairanda22,Amberg07,Furukawa08,Salzmann11a,Pons-Moll17,Xiang20a,Taylor12,Groueix18b}, known as Shape from Template (SfT) and 3D registration, assumes the availability of 3D surface templates. These methods aim to minimize the difference between the observations, such as captured images or point clouds, and the given template surface to recover the deformed state of the objects with geometric constraints or differentiable physics simulators. Unfortunately  3D templates are rarely available. Thus, there are many approaches that rely on free-form techniques without any geometric prior~\cite{Newcombe15,Bozic20a,Bozic20b,Parashar24}, which warp and accumulate different observations across frames into a 3D volume. However, these methods face challenges when reconstructing regions that remain occluded throughout the sequence. While the algorithms of~\cite{Dou16,Dou17b,Niemeyer19b,Palafox21,Li21j,Lin22} can recover full object geometry from RGB-D videos, the shape representations they use are designed for watertight surfaces, and thus not suitable for garments that are thin open surfaces.

In contrast, GarmentNets~\cite{Chi21} recovers the full 3D surfaces of previously unseen garments from point clouds. It leverages the Normalized Object Coordinate Space (NOCS) \cite{Wang19} as a category-specific canonical representation for garments. The garment mesh is reconstructed by mapping the predicted canonical mesh to the deformed one. However, GarmentNets only handles garments being grasped. To cover the much wider range of possible states that garments can be in, a large-scale garment manipulation dataset is introduced in~\cite{Xue23b}. It relies on a VR system and uses  it to learn a tracking model for estimating the complete pose of a given garment. Despite promising results, requiring a canonical geometry for garment and an initial shape estimate in the first frame imposes limitations similar to those of template-based methods. In contrast, our proposed method is not subject to these and can recover 3D meshes of garments with unknown geometry.

\parag{On-Body Garment Reconstruction.}

Clothed human reconstruction has received significant attention in recent years. However, the majority of methods primarily focus on clothing that tightly adheres to the body. In these, garments are represented either explicitly using template meshes \cite{Danerek17,Bhatnagar19,Jiang20d,Liu23b} or implicitly by signed and unsigned distance functions \cite{Corona20a,Li22c,DeLuigi23}. To handle loose-fitting garments such as skirts and dresses, the methods of~\cite{Yang18f,Zhu20,Zhu22} leverage complex physics simulation steps or feature line estimation to align the garment surface with the input image. The one of~\cite{Hong21} reconstructs garments from point clouds by predicting displacement and principal component analysis (PCA) coefficients for the mesh registered to base templates. While effective, these methods' reliance on garment templates limits their generality. To address this limitation, \cite{Li24a} uses the Implicit Sewing Patterns (ISP) model from~\cite{Li23a} as the garment representation and fits an image-conditioned deformation model to the normal estimation of the garments. 

However, all these methods rely on the articulated body shape model \cite{Loper14} as a prior. This makes them unsuitable for the perception task in the context of robot manipulations where no body is involved and garments can exhibit higher levels of crumpled deformation and occlusions, which is the focus of our work.

\parag{Diffusion Model.}

Diffusion models \cite{Ho20a,Song21} are a class of generative models that excel at learning complex data distributions through score matching. By iteratively denoising the data, these models can generate high-quality samples. They have achieved state-of-the-art performance in a wide range of image-based generative tasks \cite{Dhariwal21,Chung22a,Chung22b,Rombach22}. Additionally, diffusion models have found application in various 3D tasks, such as text-to-3D generation \cite{Xu23c,Poole22,Xu23c}, image-to-3D generation \cite{Muller23,Xu23c,Anciukevivcius23,Liu23f}, and point cloud synthesis \cite{Tyszkiewic23,Melas23}.
Recently, \cite{Guo24} introduced a diffusion-based shape prior for on-body garment registration, employing UV maps to parameterize the garment. However, its prior is specific to a single garment piece and requires coarse registration of the input point clouds. In contrast, our proposed method is capable of handling garments with diverse geometries and does not impose any registration requirements.


\section{Method}
\label{sec:method}

\begin{figure}
    \centering
    \includegraphics[width=0.99\textwidth]{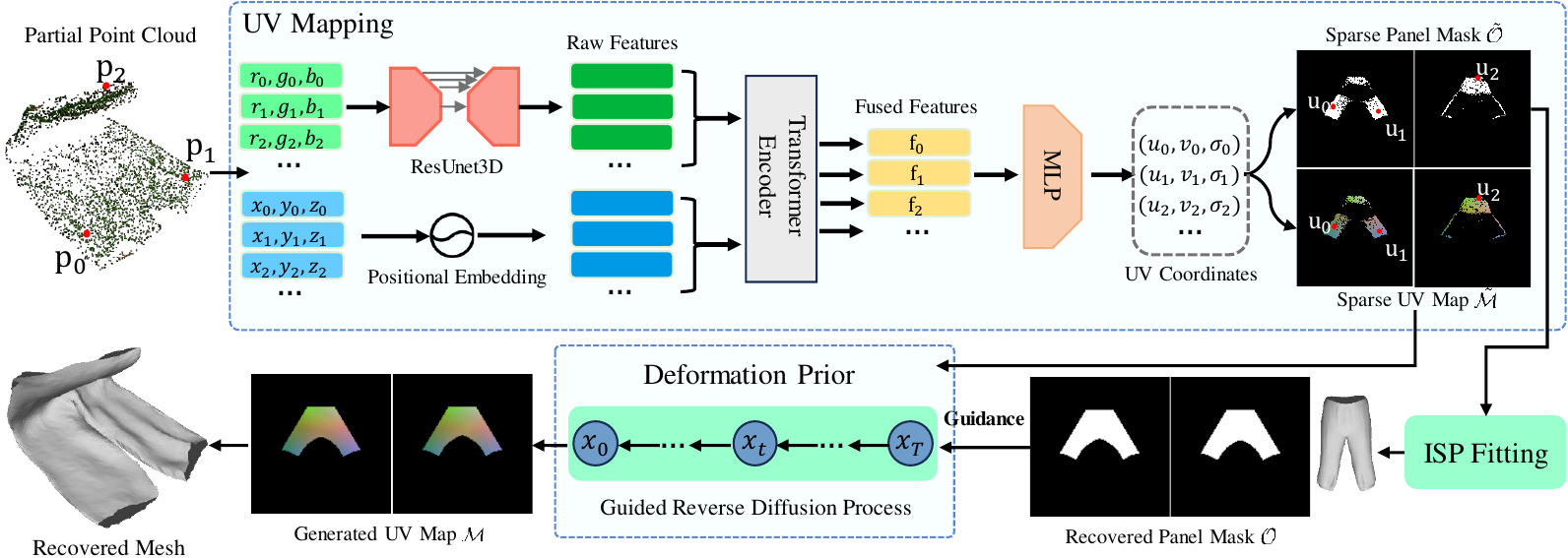}
    \caption{\textbf{Our framework}. Given a point cloud, we first map it to UV space to obtain sparse UV maps $\tilde{\mathcal{M}}$ and panel masks $\tilde{\mathcal{O}}$. We recover complete UV maps $\mathcal{M}$ and panel masks $\mathcal{O}$ from them using ISP and a deformation prior, enabling the reconstruction of the deformed garment's 3D mesh.}
    \label{fig:pip}
\end{figure}

Given a point cloud $\bP \in \mathbb{R}^{N\times3}$ that partially represents a previously unknown garment, we want to reconstruct an accurate 3D model that captures both its geometry and deformation. Instead of doing this directly in 3D space, we first use a trained UV mapper to map 3D points to the unified 2D UV space in which individual ISP~\cite{Li23a} garment panels are represented. In practice, we typically use two such panels, one for the front, and one for the back. In the resulting {\it UV maps}, each pixel can either be empty or contain the 3D location of a point. We then fit the 2D panels to these and use a reverse diffusion process to fill the potentially large holes in the UV maps, which yields a complete reconstruction. Fig.~\ref{fig:pip} depicts this process.

In this section, we first briefly describe the Implicit Sewing Pattern (ISP) garment model~\cite{Li23a}, which represents the garment geometry using one UV map for each 2D panel and which we extend by adding a diffusion-based ~\cite{Ho20a,Song21} deformation model. This makes it possible to model plausible and potentially large garment deformations. We then introduce the UV mapper that maps the point cloud data to UV space of the individual panels and discuss our approach to fitting the augmented ISP model to the resulting UV maps. 

\subsection{ISP Garment Model}
\label{sec:isp}

\parag{Formalization.} 

ISP~\cite{Li23a}  is a garment model inspired by the sewing patterns used in the fashion industry to design and manufacture clothes. Such a pattern consists of several 2D panels along with stitch information for assembling them. They are implicitly modeled using a 2D signed distance field (SDF) and a 2D label field, respectively.  For a specific garment, its corresponding latent code $\bz$, and a point $\bu$ in the 2D UV space $\Omega=[-1,1]^2$, the ISP model outputs the signed distance $s$ to the panel boundary and a label $l$ using a fully connected network $\mathcal{I}_{\Teta}$ as
\begin{equation} \label{eq:pattern}
    (s,l) = \mathcal{I}_{\Teta}(\bu,\bz) \; . 
\end{equation}
The zero crossing of the SDF defines the shape of the panel, with $s<0$ indicating that $\bu$ is within the panel and $s>0$ indicating that $\bu$ is outside the panel. The label $l$ encodes the stitch information, indicating which panel boundaries should be stitched together. To map the 2D sewing patterns to 3D surfaces, a UV parameterization function $\mathcal{A}_{\Phi}$ is learned to perform the 2D-to-3D mapping. It is written as
\begin{equation} \label{eq:uv}
    \bX = \mathcal{A}_{\Phi}(\bu,\bz) \; ,
\end{equation}
where $\bX\in\mathbb{R}^3$ represents the 3D position of $\bu$. ISP effectively registers different garments onto a unified UV space and establishes the mapping functions between points in UV space and the garments' 3D surfaces. Crucially, this is a differentiable representation. Given masks or contours of the panels, we can easily fit the latent code $\bz$ to recover the corresponding garment geometry.

\parag{Training.}
Training ISP requires the 2D sewing patterns of 3D garments in a rest state, which are not available in most garment datasets, such as CLOTH3D \cite{Bertiche20}. Following the garment flattening approach described in \cite{Li24a,Pietroni22}, we cut the garment mesh into a front and a back piece according to predefined cutting rules and then unfold them into 2D panels by minimizing an as-rigid-as-possible energy~\cite{Liu08c} to ensure local area preservation. For each garment in the dataset, we generate a front and a back panel as its sewing pattern. By pairing these 2D sewing patterns with their corresponding 3D meshes, we follow the training procedure of \cite{Li23a} to learn the weights of  the ISP model $(\mathcal{I}_{\Teta}, \mathcal{A}_{\Phi})$.

\subsection{Modeling Deformations}
\label{sec:deformation}

Although the UV parameterization described above is good at representing garments in their rest state, it does not capture the various deformations that can occur when the garment is subjected to external forces, such as folding or creasing. To address this and model the possibly large deformations of garments, we incorporate a deformation prior into ISP.

Given a set of deformed garments whose rest states are modeled by ISP, we write the corresponding UV maps as
\begin{equation}\label{eq:uv_map}
    \mathcal{M}[u,v] = 
    \begin{cases}
        \bV, & \mbox{if } s_{\bu} \le 0\\
        \varnothing, & \mbox{if } s_{\bu} > 0
    \end{cases}
    \; ,
\end{equation}
where $\bV\in \mathbb{R}^3$ is the corresponding position on the deformed mesh surface for the UV point $\bu=(u,v)$, $s_{\bu}$ is the SDF value of $\bu$, $[\cdot,\cdot]$ denotes standard array addressing and $\varnothing=(-1,-1,-1)$. Note that $\varnothing$ indicates that $\bu$ is out of the panel and has no corresponding 3D point. Each $\mathcal{M}$ represents a specific deformed state for a particular garment. To capture the distribution of plausible deformations represented in this way, we learn a deformation prior using a standard diffusion model~\cite{Ho20a,Song21}.

\parag{Diffusion.}

The popular Denoising Diffusion Probabilistic Model (DDPM) framework~\cite{Ho20a} comprises a forward and a reverse process. The forward process perturbs the clean data $\mathbf{x}_0 \sim q(\mathbf{x}_0)$ by iteratively adding Gaussian noise $\boldsymbol{\epsilon} \sim \mathcal{N}(0, \mathbf{I})$ to it. This is written as
\begin{equation}
    \mathbf{x}_{t}=\sqrt{1-\beta_{t}}\mathbf{x}_{t-1}+\sqrt{\beta_{t}} \boldsymbol{\epsilon} \; ,
\label{eq:forward}
\end{equation}
where $\mathbf{x}_t$ is the noised intermediate state at step $t = 1,2,..,T$, and $\beta_t \in (0,1)$ denotes the variance schedule. The reverse process recovers the clean data from random noise with a trained neural network $\boldsymbol{\epsilon}_{\theta}$
\begin{equation}
    \mathbf{x}_{t-1} = \frac{1}{\sqrt{\alpha_t}}\left(\mathbf{x}_t - \frac{\beta_t}{\sqrt{1-\Bar{\alpha}_t}} \boldsymbol{\epsilon}_{\theta}(\mathbf{x}_t, t)\right) + \sigma_t\mathbf{z} \; ,
    \label{eq:reverse}
\end{equation}
where $\alpha_t=1-\beta_t$, $\Bar{\alpha}_t=\prod_{i=1}^t\alpha_i$, $\mathbf{z}\sim \mathcal{N}(0, \mathbf{I})$ and $\sigma_t=\sqrt{\frac{1-\bar{\alpha}_{t-1}}{1-\bar{\alpha}_t}\beta_t}$.

\parag{Training.} 

In our context, we concatenate the UV map $\mathcal{M}$ generated by Eq. \ref{eq:uv_map} with the panel mask $\mathcal{O}$ along the channel dimension to form the training samples $\mathbf{x}_{0} =  \left[\mathcal{M}, \mathcal{O}\right]$ where
\begin{equation}\label{eq:uv_mask}
    \mathcal{O}[u,v] = 
    \begin{cases}
        1, & \mbox{if } s_{\bu} \le 0\\
        0, & \mbox{if } s_{\bu} > 0
    \end{cases}
    \; .
\end{equation}
The panel mask $\mathcal{O}$ encodes the shape of the panels as well as the 3D geometry of the canonical garment. The network $\boldsymbol{\epsilon}_{\theta}$ is trained on corrupted $\mathbf{x}_{0}$ to predict the noise by minimizing the loss
\begin{equation}
    L = \| \boldsymbol{\epsilon} - \boldsymbol{\epsilon}_{\theta}\left( \sqrt{\bar{\alpha}_{t}}\mathbf{x}_{0}+\sqrt{1-\bar{\alpha}_{t}}\boldsymbol{\epsilon}, t \right) \|_2 \; .
\end{equation}
Once the diffusion model is trained, it learns the deformation prior, enabling it to generate or recover realistic and diverse deformations for different garments.

\subsection{Mapping Point Cloud to UV Space}
\label{sec:mapping}

To relate an input 3D point cloud $\bP$ of the garment to the UV space in which the deformation  prior is learned, we rely on the UV mapper $\mathcal{G}$ shown at the top of Fig.~\ref{fig:pip}. For each 3D point, it predicts $\sigma$, the probability of belonging to  either the front or back panel, along with the $u,v$ coordinates of the pixel  where the 3D point should be stored in the UV map. As in~\cite{Xue23b}, we use a sparse 3D convolution network \cite{Choy19} to extract raw features for each point $\bp_i$ in $\bP$. These raw features are then passed through a transformer encoder with self-attention, producing fused per-point features $\mbf_i$ that capture relationships across points. These are fed to an MLP that predicts the per-point UV coordinates. It outputs probability distributions $\phi_u\in\mathbb{R}^K$ and $\phi_v\in\mathbb{R}^K$ over $K$ discrete values for the $u$- and $v$-axes, along with $\sigma$. 
$\mathcal{G}$ is trained by minimizing
\begin{equation}
    L_{\mathcal{G}} = CE(\phi_u, \hat{\phi}_u) + CE(\phi_v, \hat{\phi}_v) + BE(\sigma, \hat{\sigma}) \; ,
\end{equation} 
where $\hat{\cdot}$ denotes the ground-truth values, and $CE$ and $BE$ are the cross entropy and the binary cross entropy, respectively. 

Once trained, $\mathcal{G}$ assigns to each point $\bp_i$ a UV coordinate $\bu_i=(u_i,v_i)$ in the front ($\sigma_i\ge 0.5$) or the back ($\sigma_i < 0.5$) panel with 
\begin{equation}\label{eq:coordinate}
    u_i = -1 + \frac{2k_u}{K-1}, ~ v_i = -1 + \frac{2k_v}{K-1} \; ,
\end{equation}
where $k_u = \argmax\limits_{k\in\{0,...,K-1\}} \phi_{u}^k$ and $k_v = \argmax\limits_{k\in\{0,...,K-1\}} \phi_{v}^k$.

We then combine these predictions with $\tilde{\mathcal{M}}[u_i,v_i]=\bp_i$ and $\tilde{\mathcal{O}}[u_i,v_i]=1$ at pixels where a 3D point is projected, and $\tilde{\mathcal{M}}[u_i,v_i]=\varnothing$ and $\tilde{\mathcal{O}}[u_i,v_i]=0$ elsewhere, producing the assembled UV map $\tilde{\mathcal{M}}$ and the panel mask $\tilde{\mathcal{O}}$.

\subsection{Fitting the Model}
\label{sec:guide}

When the garment deformations are severe, there are many occlusions, and $\tilde{\mathcal{M}}$ and $\tilde{\mathcal{O}}$ are typically sparse. Nevertheless, we can use the deformation model of Section~\ref{sec:deformation} to fill-in the holes and recover complete UV maps.  To this end, we first recover the 2D shape of the 2D panels and then their individual 3D surfaces, as shown at the bottom of Fig.~\ref{fig:pip}. 

\parag{Panel Recovery.} 

To recover the 2D shape of the panels, we find the latent code $\bz$ of Eq.~\ref{eq:pattern}  that yields patterns matching $\tilde{\mathcal{O}}$ as well as possible. We take it to be
\begin{equation} \label{eq:z} 
    \bz^* = \argmin\limits_{\bz} \sum\limits_{\scalebox{.6}{$\bu\in\mathcal{O}_+$}}R(-s_\bu(\bz)) - \lambda_{area}\sum\limits_{\scalebox{.6}{$\bu\in\Omega$}}s_\bu(\bz) + \lambda_\bz||\bz||_2\; ,
\end{equation}
where $\mathcal{O}_+=\{\bu|\tilde{\mathcal{O}}_\bu=1, \bu\in\Omega\}$, $R(\cdot)$ is the ReLU function, $s_\bu(\bz)$ is the SDF value of $\bu$ computed by ISP, and $\lambda_{area}$ and $\lambda_\bz$ are the weighting constants. Since the second item of the objective function in Eq.~\ref{eq:z} penalizes large panel area, this yields panels $\mathcal{O}$ whose contours---the zero crossings of the SDF---surround the non-zero point of $\tilde{\mathcal{O}}$ as closely as possible, as shown in the bottom right of Fig.~\ref{fig:pip}.


\begin{figure}
    \centering
    \includegraphics[width=0.95\textwidth]{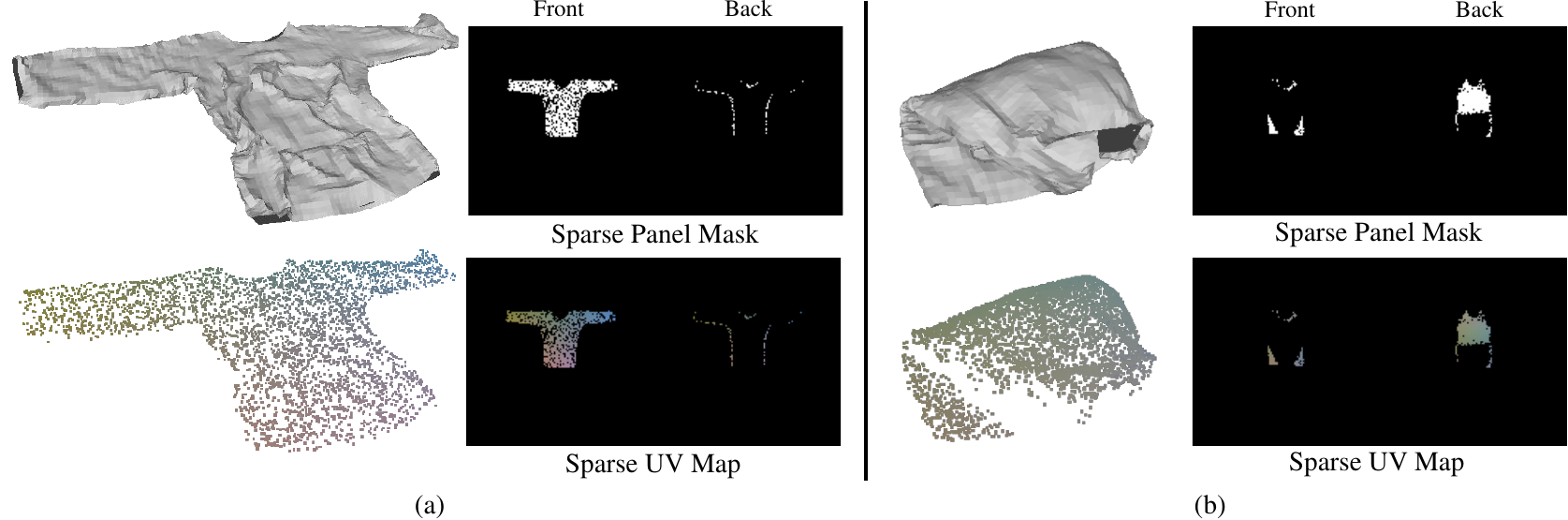}
    \caption{\textbf{The projected sparse masks $\tilde{\mathcal{O}}$ and UV maps $\tilde{\mathcal{M}}$} of the point clouds with (a) the maximum volume and (b) the minimum volume. The point clouds are color coded by their 3D positions.}
    \label{fig:cover}
\end{figure}

Given a point-cloud sequence, we solve Eq. \ref{eq:z} only once, specifically at the frame where the point cloud occupies the maximum volume in 3D space. This choice is motivated by the fact that a large volume of the point cloud indicates less deformation and occlusion of the garment, resulting in a more visible and informative mask $\tilde{\mathcal{O}}$, as shown in Fig.~\ref{fig:cover}. Using this frame for the optimization of $\bz$ leads to more accurate fitting results as demonstrated in the Appendix.


\begin{figure}[ht!]
    \centering
    \includegraphics[width=0.99\textwidth]{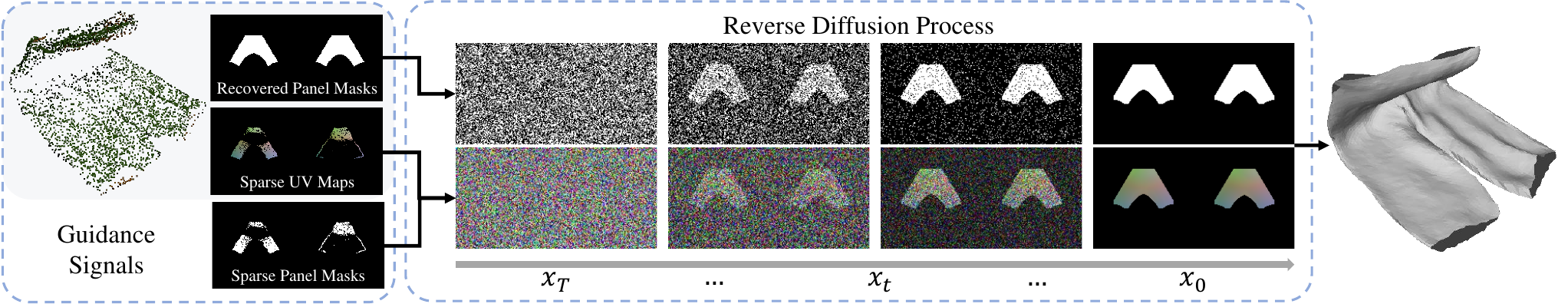}
    \caption{\textbf{The guided reverse diffusion process}. The UV maps of the deformed garment are generated by using the observations from the input point cloud as guidance to direct the reverse diffusion process.}
    \label{fig:denoise}
\end{figure}

\parag{Deformation Recovery.} 

The deformation model of Section~\ref{sec:deformation}  has been  trained to generate UV maps representing plausible deformations of garments of many different geometries. To align the generation process with the observation of the sparse UV map $\tilde{\mathcal{M}}$ and the ISP-recovered mask $\mathcal{O}$ introduced in the previous paragraph, we use them as manifold guidance~\cite{Chung22a,Chung22b} in the reverse diffusion process. We write
\begin{align}
    \nabla_{\bx_t}\log p(\bx_t|\tilde{\mathcal{M}}, \tilde{\mathcal{O}}, \mathcal{O}) &\simeq -\frac{\epsilon_\theta(\bx_t, t)}{\sigma_t}-\rho\nabla_{\bx_t}d(\hat{\bx}_0, \tilde{\mathcal{M}}, \tilde{\mathcal{O}}, \mathcal{O}) \;  \label{eq:guided_loss} \;  , \\
    \hat{\bx}_0 &= \frac{1}{\sqrt{\bar{\alpha}_t}}\bx_t - \sqrt{\frac{1-\bar{\alpha}_t}{\bar{\alpha}_t}}\epsilon_\theta(\bx_t, t)\; \label{eq:loss_uv}  \; ,
\end{align} 
where $\rho$ is the guidance step size. The function $d$ measures the difference between the generated result and the observations
\begin{equation} \label{eq:guidance} 
    d(\hat{\bx}_0, \tilde{\mathcal{M}}, \tilde{\mathcal{O}}, \mathcal{O}) = \lVert\tilde{\mathcal{O}}*(\hat{\bx}_{0,\mathcal{M}} - \tilde{\mathcal{M}})\rVert _2 + \lVert\hat{\bx}_{0,\mathcal{O}} - \mathcal{O}\rVert _1 \; ,
\end{equation}
where $\hat{\bx}_{0,\mathcal{M}}$ and $\hat{\bx}_{0,\mathcal{O}}$ refer to the generated UV map and panel mask respectively, and $*$ denotes the elementwise multiplication. When sequential information is available, we additionally refine our method by using the previous prediction $\mathcal{M}_{prev}$ as the regularization for the unobserved part
\begin{equation} \label{eq:guidance_t} 
    d(\hat{\bx}_0, \tilde{\mathcal{M}}, \tilde{\mathcal{O}}, \mathcal{O}) = \lVert\tilde{\mathcal{O}}*(\hat{\bx}_{0,\mathcal{M}} - \tilde{\mathcal{M}})\rVert _2 + \lVert\hat{\bx}_{0,\mathcal{O}} - \mathcal{O}\rVert _1 + \lambda \lVert(1-\tilde{\mathcal{O}})*(\hat{\bx}_{0,\mathcal{M}} - \mathcal{M}_{prev})\rVert _2.
\end{equation}
where $\lambda$ is a weighting constant. Finally, the garment mesh is inferred from the generated UV map using the mapping function of ISP. As illustrated in Fig. \ref{fig:denoise}, this process finally produces a garment mesh that aligns with the point cloud observation.


\section{Experiments}

\subsection{Dataset, Evaluation Metrics, and Baseline}
\label{sec:setting}
We train our models using data from the VR-Folding \cite{Xue23b} and CLOTH3D \cite{Bertiche20} datasets. The VR-Folding dataset is collected using a VR-based recording system, where participants manipulate garments (i.e., folding and flattening) in a simulator through a VR interface. The dataset features pants, shirts, tops, and skirts selected from the CLOTH3D, and each category covers a wide shape range. VR-Folding comprises 9767 manipulation videos and 790K multi-view RGB-D frames, which are used to generate point clouds. For the training of ISP, we use the same garments from CLOTH3D as those selected in VR-Folding and generate their sewing patterns as described in Sec. \ref{sec:isp}. We generate UV maps of deformed garments and the corresponding UV coordinates of point clouds using the mapping function of ISP. These UV maps and coordinates serve to train the diffusion model and the UV mapper, respectively. For each garment category, we train a separate set of models, using the same training and test splits as \cite{Xue23b}.

As in~\cite{Xue23b,Chi21}, we employ the Chamfer Distance $D_{cf}$ and the Correspondence Distance ($D_{cr}$, $A_d$) as evaluation metrics. $D_{cf}$ measures the surface reconstruction quality by calculating the Chamfer distance in centimeters from the reconstructed mesh to the ground truth. $D_{cr}$ represents the point-wise L2 distance in centimeters between the reconstruction and the ground truth, which evaluates the accuracy of garment pose estimation. Note that the correspondences of $D_{cr}$ are established by finding the closest point of the ground truth in canonical space instead of the deformed one as $D_{cf}$. Finally, we take $A_d$ to be the ratio of frames with $D_{cr}$ < $d$ cm.

We compare our method against state-of-the-art approaches: GarmentNets \cite{Chi21} and GarmentTracking \cite{Xue23b}. GarmentTracking estimates the per-vertex garment pose based on the given canonical garment mesh and the initialization of the first frame. GarmentNets is a single-frame garment shape estimation method that utilizes the winding number field for garment meshing. Like our method, GarmentNets does not require ground truth geometry.

\subsection{Quantitative Results}
\label{sec:quant}
\begin{table}[h!]
    \begin{center}
      \scalebox{.85}{
      \begin{tabular}{c|c|c|ccc|c|ccc|c}
        \toprule
        \multicolumn{1}{c|}{\multirow{2}{*}{Type}} & \multicolumn{1}{c|}{\multirow{2}{*}{Method}} & \multicolumn{1}{c|}{\multirow{2}{*}{Init.}} & \multicolumn{4}{c|}{Folding} & \multicolumn{4}{c}{Flattening}\\ \cline{4-11}
        & & & $A_{3}\uparrow$ & $A_{5}\uparrow$ & $D_{cr}\downarrow$ & $D_{cf}\downarrow$ & $A_{5}\uparrow$ & $A_{10}\uparrow$ & $D_{cr}\downarrow$ & $D_{cf}\downarrow$ \\
        \midrule
        \multirow{5}{*}{Shirt} & GarmentNets \cite{Chi21} & N/A & 0.8\% & 21.5\% & 6.40 & 1.58 & 13.2\% & 59.4\% & 10.54 & 3.54 \\ 
        &  GarmentTracking \cite{Xue23b} & GT & 29.8\% & 85.8\% & 3.88 & 1.16 & 30.7\% & 83.4\% & 8.63 & 1.78  \\ 
        &  GarmentTracking \cite{Xue23b} & Pert. & 29.0\% & 85.9\% & 3.88 & 1.18 & 25.4\% & 81.6\% & 8.94 & 1.85  \\  
        &  GarmentTracking \cite{Xue23b} & GN. & 25.4\% & 78.9\% & 4.04 & 1.18 & - & - & - & -  \\  
        &  Ours & N/A & \textbf{84.7\%} & \textbf{97.9\%} & \textbf{2.36} & \textbf{0.77} & \textbf{78.8\%} & \textbf{95.2\%} & \textbf{4.19} & \textbf{1.08} \\ 
        \midrule
        \multirow{5}{*}{Pants} & GarmentNets \cite{Chi21} & N/A & 16.2\% & 69.5\% & 4.43 & 1.30 & 1.5\% & 42.4\% & 12.54 & 4.19 \\ 
        &  GarmentTracking \cite{Xue23b} & GT & 47.3\% & 94.0\% & 3.26 & 1.07 & 31.3\% & 78.2\% & 8.97 & 1.64 \\ 
        &  GarmentTracking \cite{Xue23b} & Pert. & 42.8\% & 93.6\% & 3.35 & 1.10 & 30.7\% & 76.9\% & 9.55 & 2.71  \\ 
        &  GarmentTracking \cite{Xue23b} & GN & 45.1\% & 92.2\% & 3.33 & 1.16 & - & - & - & -  \\ 
        &  Ours & N/A & \textbf{75.9\%} & \textbf{97.9\%} & \textbf{2.69} & \textbf{0.70} & \textbf{60.8\%} & \textbf{91.4\%} & \textbf{5.32} & \textbf{1.16} \\ 
        \midrule
        \multirow{5}{*}{Top} & GarmentNets \cite{Chi21} & N/A & 10.3\% & 53.8\% & 5.19 & 1.51 &13.1\% & 42.5\% & 12.11 & 2.85 \\ 
        &  GarmentTracking \cite{Xue23b} & GT & 37.9\% & 85.9\% & 3.75 & 0.99 & 54.6\% & 82.8\% & 6.59 & 1.15  \\ 
        &  GarmentTracking \cite{Xue23b} & Pert. &36.6\% & 86.1\% & 3.76 & 1.00 & 54.2\% & 82.6\% & 7.80 & 2.59  \\
        &  GarmentTracking \cite{Xue23b} & GN &21.1\% & 61.9\% & 4.82 & 1.11 & - & - & - & -  \\
        &  Ours & N/A & \textbf{71.2\%} & \textbf{93.5\%} & \textbf{2.65} & \textbf{0.74} & \textbf{70.2\%} & \textbf{86.2\%} & \textbf{5.24} & \textbf{1.08} \\ 
        \midrule
        \multirow{5}{*}{Skirt} & GarmentNets \cite{Chi21} & N/A & 1.1\% & 30.3\% & 6.95 & 1.89 & 0.1\% & 7.9\% & 18.48 & 5.99 \\ 
        &  GarmentTracking \cite{Xue23b} & GT & 23.5\% & 71.3\% & \textbf{4.61} & 1.33 & \textbf{5.4\%} & \textbf{39.4\%} & 16.09 & 2.02  \\ 
        &  GarmentTracking \cite{Xue23b} & Pert. & 22.8\% & 70.6\% & 4.72 & 1.36 & 2.3\% & 35.5\% & 16.55 & 2.15 \\ 
        &  GarmentTracking \cite{Xue23b} & GN & 14.7\% & 65.9\% & 5.36 & 1.46 & - & - & - & - \\ 
        &  Ours & N/A & \textbf{32.5\%} & \textbf{76.5\%} & 4.70 & \textbf{1.04} & 5.1\% & 33.1\% & \textbf{14.26} & \textbf{1.75} \\ 
        \bottomrule
      \end{tabular}
    }
  \end{center}
      \caption{\textbf{Quantitative comparisons} of our method to GarmentNets and GarmentTracking on VR-Folding dataset.}
      \label{tab:quantitative}
\end{table}

Table \ref{tab:quantitative} shows the quantitative results obtained on the VR-Folding dataset. In the third column, the abbreviation "N/A" indicates the absence of any initialization, while "GT", "Pert." and "GN" represent the results of GarmentTracking using the ground-truth mesh, the noise-perturbed ground-truth mesh, and the estimation of GarmentNets as the initialization, respectively. Our method outperforms the baselines by a large margin for both the Folding and Flattening sets. GarmentNets can handle garments without prior knowledge of their geometry but has the lowest reconstruction accuracy. GarmentTracking is more accurate and benefits from using the given garment mesh as a prior, but its performance is highly influenced by the choice of initialization. When using the network-predicted result (GN) instead of the ground truth, its performance drops substantially, particularly for Shirt, Skirt and Top. This greatly restricts its applicability in real-world scenarios where obtaining the ground truth garment mesh in advance is rarely possible. In contrast, our method has no such limitation while still achieving the highest reconstruction accuracy. The performance disparity between our method and the baselines is particularly significant for challenging metrics such as $A_{3}$ (on Folding) and $A_{5}$ (on Flattening), for instance, 84.7\% vs 29.8\% and 78.0\% vs 24.5\% on Shirt. 

We also notice that both our method and the baseline models achieve relatively higher $D_{cr}$ and lower $A_{d}$ values for Skirt compared to other categories. This discrepancy arises from the ambiguous definition of skirt sides due to its rotational symmetry. When the skirt is rotated by a specific amount around the medial axis, the resulting shape is nearly identical to the original one, which can yield high $D_{cr}$ and low $A_{d}$ values because they are computed using correspondence between the estimated canonical mesh and the ground truth. Consequently, this symmetrical ambiguity makes these metrics unsuitable for assessing the reconstruction quality of skirts, whereas the Chamfer distance $D_{cf}$ does not have this issue, on which we obtain the lowest values. An illustrative example is provided in the Appendix.


\begin{figure}[th!]
    \centering
    \includegraphics[width=0.99\textwidth]{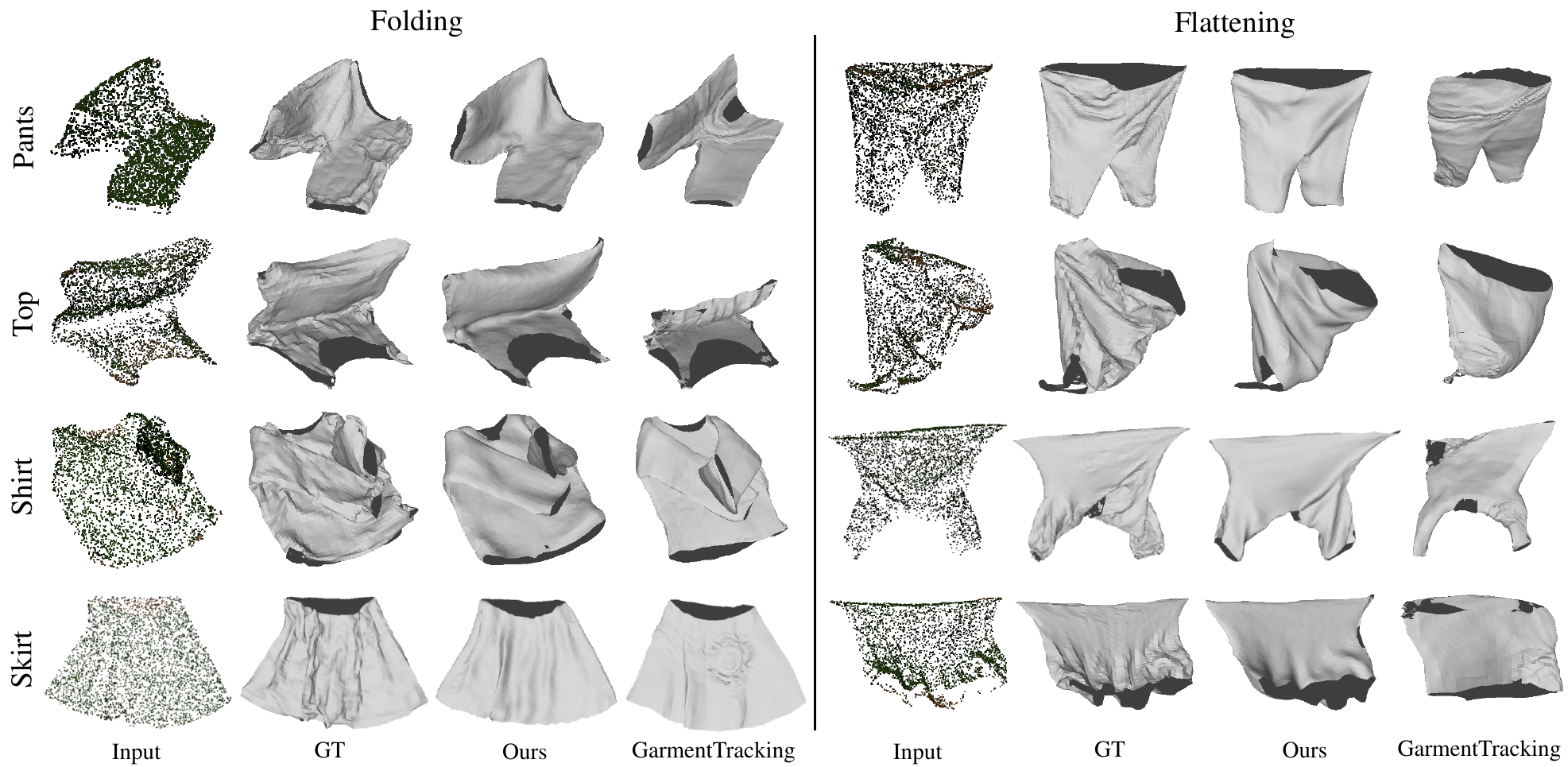}
    \caption{\textbf{Qualitative comparisons} of our method to GarmentTracking (initialized with ground truth meshes) on VR-Folding dataset for the categories of Pants, Top, Shirt and Skirt. }
    \label{fig:fold}
\end{figure} 


\begin{figure}[ht!]
    \centering
    \includegraphics[width=0.99\textwidth]{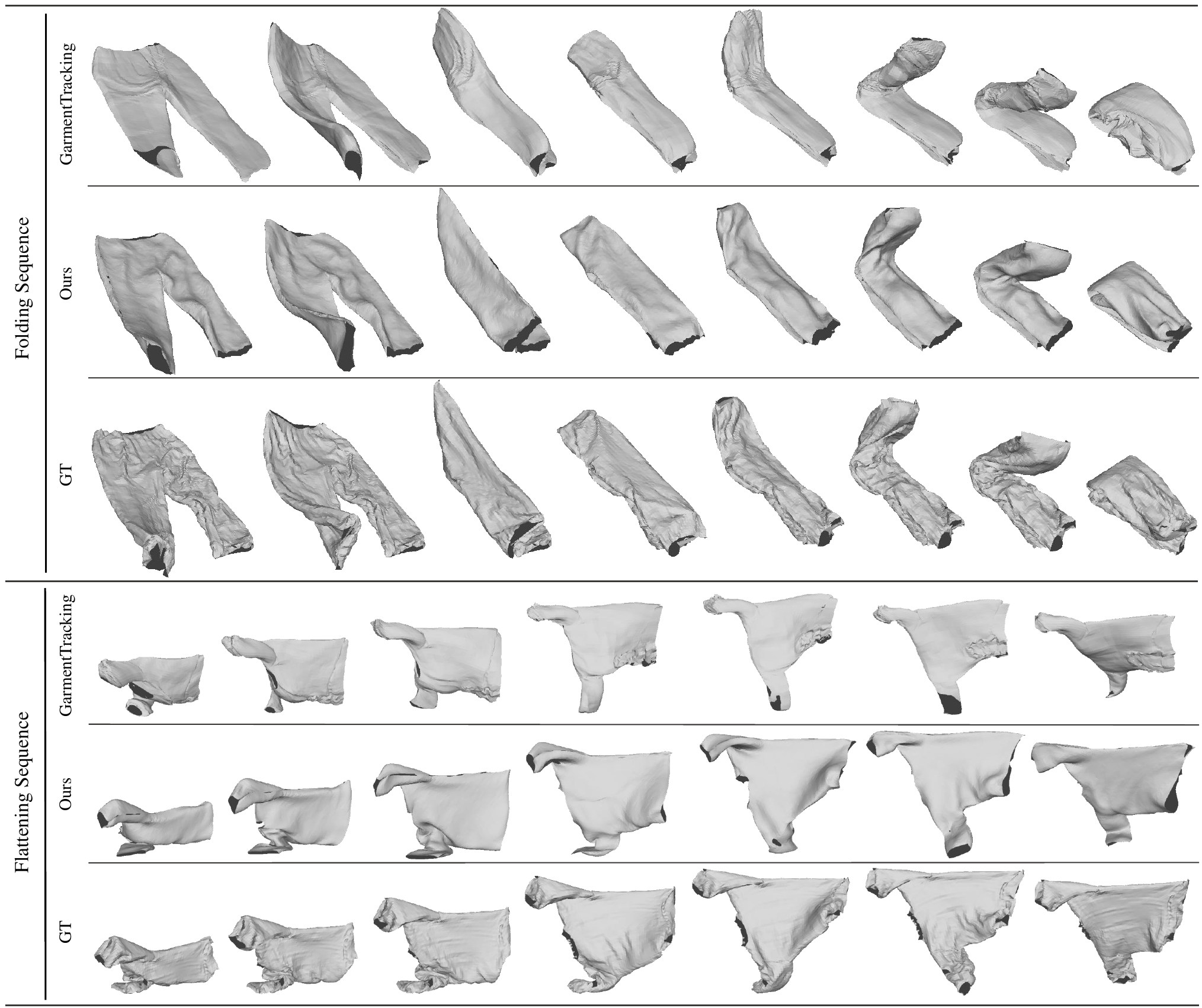}
    \caption{\textbf{Qualitative comparisons} of our method to GarmentTracking (initialized with ground truth meshes) on VR-Folding dataset for the sequences of Folding and Flattening.}
    \label{fig:seq_fold}
\end{figure} 

\subsection{Qualitative Results}
In Fig. \ref{fig:fold}, we show the qualitative comparison with GarmentTracking which uses the ground truth garment mesh as the initialization. GarmentTracking produces results with inaccurate size and deformation, and unrealistic artifact can show up on the reconstructed surfaces. In contrast, our method can recover garment meshes from input point clouds faithfully with correct shape and deformations. In Fig. \ref{fig:seq_fold}, we further show the reconstructed results for a folding and a flattening sequences, which demonstrates our method can consistently produce accurate results compared with GarmentTracking. More qualitative comparisons can be found in the Appendix.

\begin{figure}[ht!]
    \centering
    \includegraphics[width=0.95\textwidth]{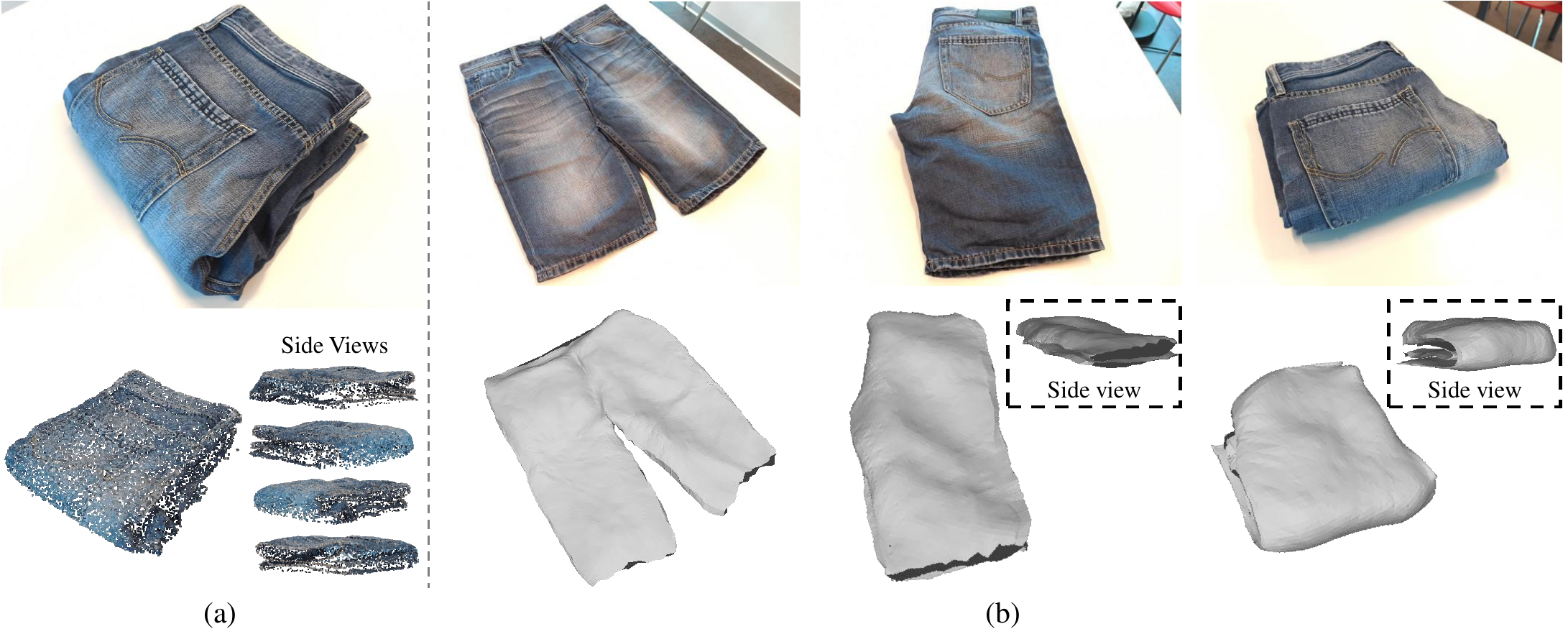}
    \caption{\textbf{Real-world evaluation}. (a) The captured image and point cloud of the pants. (b) Our reconstructed results.}
    \label{fig:real_pants}
\end{figure} 

\subsection{Evaluation on Real-World Data}
To evaluate our method on real-world data, we capture RGB images of a pair of pants and a sweater, and compute dense point clouds using the \textit{nerfstudio} library \cite{Tancik23} for them, as illustrated in Fig.~\ref{fig:real_pants} (a). We remove background points and use the resulting point cloud downsampled as input to our method. Fig.~\ref{fig:real_pants} (b) shows the qualitative results for the pants (the results of the sweater are included in the Appendix). Despite being trained on simulated data, our method is able to reconstruct 3D meshes for both flat and folded garments in real-world scenarios.


\section{Conclusion}

We have proposed a method that addresses the challenges of reconstructing garment that is not being worn and can be manipulated in complex ways. It leverages the Implicit Sewing Patterns (ISP) model for geometry modeling, a generative diffusion model for learning deformation prior, and a UV mapping network to relate the 3D point cloud observations to the UV space where the priors are learned. We have demonstrated the effectiveness of our fitting approach in accurately reconstructing garment meshes in the presence of severe self-occlusion and unknown garment geometries. In future work, we will incorporate accumulated point-cloud information across time to improve the accuracy of UV mapping and mesh reconstruction. 

\parag{Acknowledgement.} This project was supported in part by the Swiss National Science Foundation.


\bibliographystyle{unsrt} 
\bibliography{string,geom,graphics,learning,vision,optim,robotics,misc,biomed}

\end{document}